\documentclass{article}
\usepackage{spconf}

\usepackage{enumitem}
\usepackage{cite}
\usepackage{amsmath,amssymb,amsfonts}
\usepackage{algorithmic}
\usepackage{graphicx}
\usepackage{textcomp}
\usepackage{xcolor}
\usepackage{array}
\usepackage{mathtools}
\usepackage{tcolorbox}
\usepackage{color}
\usepackage{theorem}
\usepackage{amssymb}
\usepackage{caption}
\usepackage{subcaption}
\usepackage{cite,hyperref}
\usepackage{cases}
\usepackage{url}
\usepackage{algorithmic}
\usepackage{algorithm}
\usepackage{tikz}
\usetikzlibrary{shapes,arrows}
\input{my_styles.sty}
\usepackage{multirow}
\usepackage{hhline}

\newtcolorbox{myblockt}[1]{colback=urblue!5!white,
	colframe=urblue,fonttitle=\bfseries,
	title=#1}
\newtcolorbox{myblock}{colback=urblue!5!white,
	colframe=urblue,fonttitle=\bfseries}
\def\BibTeX{{\rm B\kern-.05em{\sc i\kern-.025em b}\kern-.08em
    T\kern-.1667em\lower.7ex\hbox{E}\kern-.125emX}}

\begin{document}
\ninept

\title{Recovering Missing Node Features with Local Structure-based Embeddings}
\name{Victor M. Tenorio$^*$, Madeline Navarro$^\dagger$, Santiago Segarra$^\dagger$, and Antonio G. Marques$^*$ 
\thanks{This work was partially supported by the NSF under award CCF-2008555, by the Spanish AEI (AEI/10.13039/ 501100011033) grants PID2019-105032GB-I00, PID2022-136887NB-I00 and FPU20/05554, by the Young Researchers R\&D Project, ref. num. F861 (CAM and URJC), and by the Comunidad de Madrid (Madrid ELLIS Unit). Research was sponsored by the Army Research Office and was accomplished under Grant Number W911NF-17-S-0002. The views and conclusions contained in this document are those of the authors and should not be interpreted as representing the official policies, either expressed or implied, of the Army Research Office or the U.S. Army or the U.S. Government. The U.S. Government is authorized to reproduce and distribute reprints for Government purposes notwithstanding any copyright notation herein.
Emails: \href{mailto:victor.tenorio@urjc.es}{victor.tenorio@urjc.es}, \href{mailto:nav@rice.edu}{nav@rice.edu},  \href{mailto:segarra@rice.edu}{segarra@rice.edu}, \href{mailto:antonio.garcia.marques@urjc.es}{antonio.garcia.marques@urjc.es}}}
\address{$^*$ Dept. of Signal Theory and Communications, King Juan Carlos University, Madrid, Spain \\
$^\dagger$ Dept. of Electrical and Computer Engineering, Rice University, Houston, USA}
\maketitle






\begin{abstract}
Node features bolster graph-based learning when exploited jointly with network structure.
However, a lack of nodal attributes is prevalent in graph data.
We present a framework to \emph{recover completely missing node features for a set of graphs}, where we only know the signals of a subset of graphs.
Our approach incorporates prior information from both graph topology and existing nodal values.
We demonstrate an example implementation of our framework where we assume that node features depend on local graph structure.
Missing nodal values are estimated by aggregating known features from the most similar nodes.
Similarity is measured through a node embedding space that preserves local topological features, which we train using a Graph AutoEncoder.
We empirically show not only the accuracy of our feature estimation approach but also its value for downstream graph classification.
Our success embarks on and implies the need to emphasize the relationship between node features and graph structure in graph-based learning.
\end{abstract}


\begin{figure}[!t]
    \centering
    \includegraphics[width=0.3\textwidth]{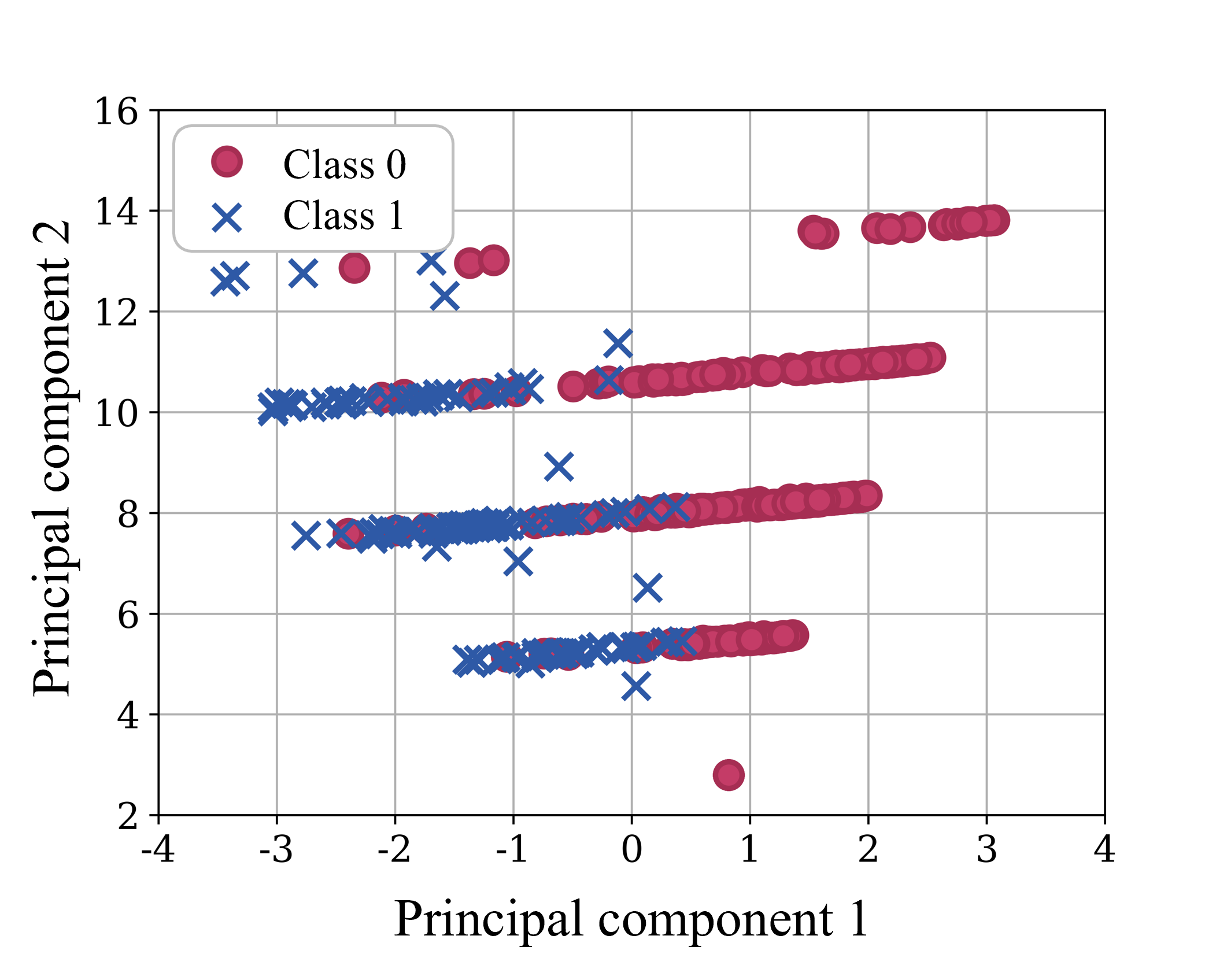}
    \caption{\small{Embeddings for nodes in graphs from the AIDS molecule dataset.
    Each point is a node embedding based on local structural characteristics, such as degree.
    Nodes corresponding to graphs of different classes are shifted in the embedding space, implying that local structure is correlated with molecule class for the AIDS dataset.}}
    \label{fig:node_embeddings}
    \vspace{-0.5cm}
\end{figure}

\section{Introduction}

\begin{figure*}[t]
    \centering
    \includegraphics[width=.8\textwidth]{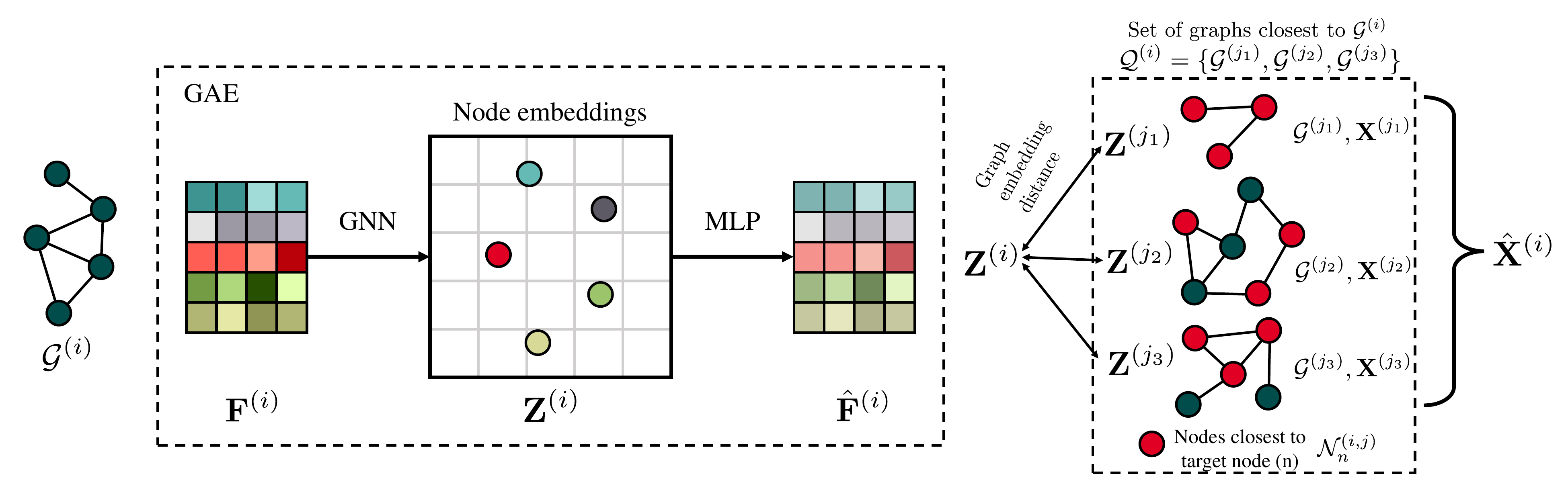}
    \caption{\small{Schematic of the proposed methodology. First, we compute a feature matrix $\bbF^{(i)}$ for each graph $\ccalG^{(i)}$ based on structural characteristics. 
    Then, we train a GAE on $\bbF^{(i)}$ to produce node embeddings $\bbZ^{(i)}$ and corresponding graph embeddings.
    The latter are used to measure similarity among graphs, and find the most similar graphs to $\ccalG^{(i)}$, collected in $\ccalQ^{(i)}$.
    The node embeddings are used to find, among the nodes of the graphs in $\ccalQ^{(i)}$, those nodes which are the closest to each node $n$ from $\ccalG^{(i)}$, collected in $\ccalN^{(i,j)}_n$.
    Finally, we estimate node features in $\ccalG^{(i)}$ by averaging the features of the closest graphs and their nodes, resulting in realistic yet accurate node feature estimates.}
    }
    \label{fig:methodology}
    \vspace{-0.5cm}
\end{figure*}

For practical applications in chemistry~\cite{gilmer2017neural,duvenaud2015convolutional}, medicine~\cite{ktenaDistanceMetricLearning2017}, and many others~\cite{KIM201886}, data can be naturally represented as interconnected entities using graphs.
Supervised learning on graphs aims to predict characteristics using both graph structure and, in some cases, node features, also known as graph signals.
These features can improve graph-based predictions when jointly used with graphs, not only when the nodal values are semantically relevant but also when the observations on nodes and the graph structure are dependent~\cite{cui22positional}.

The relationship between node features and their underlying graph is well-studied for node-level tasks.
These tasks typically entail predicting nodal characteristics for a single graph where a subset of features is known.
Classifying nodes in the semi-supervised learning setting requires the influence of the underlying graph on nodal values to propagate known information to unlabeled nodes~\cite{kipf17gnns}.
Graph signal reconstruction is a common task in graph signal processing in which partially observed features and a known graph signal model are applied together for downstream tasks, such as approximating hidden node values~\cite{chen15discrete,ramirezGraphsignalReconstructionBlind2021,Romero2016KernelBasedRO}.
In these cases, a portion of the features is known, even if a minority, and the graph topology informs how existing values provide information about those hidden.

The task of recovering completely missing graph signals for a given graph is far less explored.
As nodal values are critical for predictions and necessary for the implementation of graph neural networks (GNNs), we require a method to accurately estimate unknown node features~\cite{cui22positional, caiSimpleEffectiveBaseline2018}.
Unlike node-level predictions for which we have partial nodal observations, we cannot use the underlying graph structure to propagate existing information and infer missing values~\cite{kipf17gnns, chenLearningAttributeMissingGraphs2022}.
In such cases, the graph may belong to a family of graphs whose structural and nodal characteristics are related.
Many graph-level tasks consist of such data, such as predicting molecular structures and identifying characteristics of social networks~\cite{duvenaud2015convolutional,KIM201886}.
Existing works often characterize graph families by shared random graph models~\cite{navarroJointNetworkTopology2022} or latent embedding spaces~\cite{kipf17gnns}.
However, these works enforce a global relationship between the graphs and their signals, requiring knowledge of the entire graph.

Even under the assumption of a shared graph family, the relationship between each of the graphs and its node features is largely decoupled for graph-level learning tasks.
For example, methods that interpolate between labeled graphs for improving classification typically treat graphs and node features separately~\cite{han2022gmixup,ma2023graph,navarro23graphmad}.
Approaches that do aim to replace missing graph signals typically rely on values that possess solely topological features with no additional nodal information~\cite{cui22positional,xu2018gnns}, and many use unrelated values that do not incorporate structure~\cite{abboud21random,sato2021random}.
We empirically show that such approaches are suboptimal, even when we include structural information.

Given a set of graphs where only a subset has known node features, we present a framework to recover completely missing node features.
In particular, we propose \emph{node feature recovery for graph-level tasks incorporating both graph topology and known feature values}.
We demonstrate an implementation of this approach assuming that node features depend on local graph structural characteristics, differently from previous approaches that largely rely on aggregating information from a node's neighborhood, for example via low-pass graph filters~\cite{isufi2022graph,liu2023blind}.
Thus, we train a Graph AutoEncoder (GAE) to learn a node embedding space that preserves the local structural characteristics for each node, visualized in Fig.~\ref{fig:node_embeddings}.
In this setting, feature values are assumed to be closer when node neighborhoods are similar, so nodes with known features can effectively provide the most realistic feature estimates for those with similar local topologies.

Our contributions are as follows.
\begin{itemize}[labelwidth=1em,leftmargin =\dimexpr\labelwidth+\labelsep\relax]
\i[(i)] We present an approach to learn completely missing node features whose values are assumed to be dependent on graph structure, and we exhibit the approach in practice through the setting where features depend on local structure.
\i[(ii)] We demonstrate that for many graph classification benchmark datasets, local node structure is indeed indicative of class. 
\i[(iii)] We empirically validate the ability of our method to not only accurately learn missing node features using a set of graphs with known features, but we also demonstrate the value of recovering accurate node features for downstream tasks.
\end{itemize}

\begin{table*}[t]
    \footnotesize
    \centering
    \begin{tabular}{c|ccccccccc}
     & MUTAG & AIDS & PROTEINS & ENZYMES & ogbg-molbbbp & ogbg-molbace \\ \hline
    Zeros & $1.00\pm0.00$ & $1.00\pm0.00$ & $1.00\pm0.00$ & $1.00\pm0.00$ & $1.00\pm0.00$ & $1.00\pm0.00$ \\
    Ones & $2.450\pm0.00$ & $6.083\pm0.00$ & $1.414\pm0.00$ & $1.414\pm0.00$ & $0.806\pm0.00$ & $0.803\pm0.00$ \\
    Random & $1.510\pm0.004$ & $3.550\pm0.003$ & $0.965\pm0.002$ & $0.961\pm0.003$ & $0.896\pm0.000$ & $0.895\pm0.000$ \\
    Degree & $0.938\pm0.001$ & $1.414\pm0.006$ & $0.899\pm0.003$ & $0.891\pm0.001$ & $0.975\pm0.000$ & $0.984\pm0.000$ \\
    LSE-NG ($\bar{Q} = 1$) & $0.631\pm0.040$ & $0.790\pm0.005$ & $0.739\pm0.010$ & $0.715\pm0.005$ & $0.288\pm0.002$ & $0.264\pm0.002$ \\
    LSE-NG ($\bar{Q} = 3$) & $0.607\pm0.012$ & $0.746\pm0.005$ & $0.719\pm0.006$ & $0.716\pm0.007$ & $0.251\pm0.003$ & $0.228\pm0.001$ \\ \hline
    LSE-NN ($\bar{Q} = 1$) & $\mathbf{0.119\pm0.017}$ & $\mathbf{0.555\pm0.005}$ & $0.708\pm0.010$ & $\mathbf{0.654\pm0.013}$ & $0.177\pm0.002$ & $\mathbf{0.113\pm0.001}$ \\
    LSE-NN ($\bar{Q} = 3$) & $0.128\pm0.024$ & $0.562\pm0.005$ & $\mathbf{0.696\pm0.007}$ & $0.664\pm0.011$ & $\mathbf{0.175\pm0.004}$ & $0.126\pm0.002$
\vspace{-0.2cm}
    \end{tabular}
    \caption{\small{Normalized feature generation error $\|\bbX-\hat{\bbX}\|_F^2/\|\bbX\|_F^2$. For LSE-NG and LSE-NN, we predict missing node features for each graph from the $\bar{Q}$ nearest graphs in the dataset with respect to the graph embeddings belonging to the same class. The top performing methods are \textbf{bolded}.}} 
    \label{tab:results_npred}
    \vspace{-0.5cm}
\end{table*}

\vspace{-0.35cm}

\section{Background}

In this section, we provide the necessary background on graph-based learning, along with a review of existing approaches on node representation learning and addressing missing node features. 
We start with some basic notation. 
A graph $\ccalG = \{\ccalV, \ccalE\}$ comprises a set of nodes $\ccalV = \{1, \ldots, N\}$ and a set of edges $\ccalE = \{(n_1,n_2) | n_1,n_2 \in \ccalV\}$. 
Graphs can be conveniently represented by the so-called adjacency matrix $\bbA$. 
For edges in $\ccalG$, $(n_1,n_2)\in\ccalE$ iff $A_{n_1,n_2}=1$.
In machine learning and signal processing setups, data is often associated with each of the nodes. 
In particular, let $\bbX\in\reals^{N\times F}$ be a data matrix, whose entry $X_{n,f}$ represents the value of feature (signal) $f$ at node $n$. 
The $n$th row of $\bbX$ is typically referred to as the data features associated with node $n$ and the $f$th column of $\bbX$ as the $f$th graph signal. 
On top of these \emph{node features}, one can also associate a number of \emph{topological features}, such as centrality values or clustering coefficients, with each of the nodes~\cite{guo20role}.  




\vspace{0.1cm}

\noindent \textbf{Node representation learning.} 
Learning node representations (embeddings) has been a prevalent topic of research in the GSP literature almost since its inception~\cite{cai2017embedding,goyal18embedding}. 
Since the adjacency matrix $\bbA$ is an alternative representation of $\ccalG$, each node can be (perfectly) represented in an $N$-dimensional vector space using the corresponding row of $\bbA$. 
As a result, node representation algorithms typically aim to learn representations in a lower-dimensional space, that is, they aim at learning a matrix $\bbZ\in\reals^{N\times P}$ with $P<N$. 
The ultimate goal when designing $\bbZ$ is to sufficiently characterize nodal behavior in the context of the graph application at hand. 
Countless approaches to learn nodal representations include algorithms from random walks~\cite{perozzi14deepwalk} to GNNs~\cite{kipf17gnns,xu2018gnns}.

Most of these approaches learn node embeddings based on node proximity: the closer nodes $n$ and $n'$ are in the graph, the more similar their embeddings $\bbz_n$ and $\bbz_{n'}$ are.
Recent works have begun to emphasize learning node embeddings based on the role of each node in the graph, guided by its topological features~\cite{guo20role,cui22positional}. 
Under this setting, we may transform structural similarities between nodes into geometric relationships in the embedding space.
Inspired by this concept, we draw on such structure-based embeddings to identify which nodes are similar for sharing feature values.

\vspace{0.1cm}

\noindent \textbf{Missing node features.}
Previous works dealing with missing feature data consider partially missing node features, where only some entries of the feature matrix $\bbX$ are observed.
In such formulations, the task, known as feature imputation~\cite{rossi2021fp,you2020handling,spinelli20missing} or graph signal interpolation~\cite{chen15discrete,ramirezGraphsignalReconstructionBlind2021,Romero2016KernelBasedRO}, is to learn the missing entries of $\bbX$.
The full matrix $\hbX$ (which contains now both the given and estimated values) is then applied to a downstream task, usually for node-level tasks on a single graph, such as node classification. 
Feature imputation has been approached by graph spectral approaches~\cite{chen15discrete}, kernel approaches~\cite{Romero2016KernelBasedRO}, propagating the known features~\cite{rossi2021fp,ramirezGraphsignalReconstructionBlind2021} or using GNNs~\cite{you2020handling,spinelli20missing}.
Differently from these works, we aim to solve a more difficult graph-level version of this problem, where for a subset of (or all the) features, we do not have access to any of the nodes, and we must infer the entire set of nodal values (either multiple columns of $\bbX$ or the full matrix itself) from data associated with other graphs. 
As detailed in Sec. \ref{sec:methodology}, this paper considers the case of having access to a set of graphs, a subset of which have no observed node features (that is, no access to any of the values of $\bbX$), which is common in social networks, for example~\cite{caiSimpleEffectiveBaseline2018}.

In the setting of completely missing node features, other works replace these features with carefully crafted random matrices~\cite{abboud21random,sato2021random}, position-dependent values~\cite{you2019positionaware,li2020distance,youIdentityawareGraphNeural2021}, or structural properties, such as the degree~\cite{cui22positional,xu2018gnns}, surprisingly showing that GNNs are still able to obtain great performance without meaningful node features and only using the graph structure. 
However, random and constant features are independent of class labels, and we empirically demonstrate that using random or structural node features is suboptimal.
Moreover, our approach using a realistic estimate of node features exhibits superior performance.


\begin{table*}[t]
    \footnotesize
    \centering
    \begin{tabular}{c|ccccccccc}
         & MUTAG & AIDS & PROTEINS & ENZYMES & ogbg-molbbbp & ogbg-molbace \\ \hline
         True Features & 83.25 $\pm$ 9.52 & 97.00 $\pm$ 1.49 & 72.74 $\pm$ 3.75 & 36.17 $\pm$ 4.97 & 85.53 $\pm$ 2.38 & 74.38 $\pm$ 3.49 \\ \hline
Zeros & 78.75 $\pm$ 9.34 & 95.92 $\pm$ 1.87 & 70.09 $\pm$ 5.25 & 25.50 $\pm$ 7.15 & 77.15 $\pm$ 2.63 & 53.33 $\pm$ 5.25 \\
Ones & 82.25 $\pm$ 9.55 & 94.45 $\pm$ 1.86 & 69.25 $\pm$ 5.26 & 24.08 $\pm$ 5.10 & 77.77 $\pm$ 3.26 & 53.55 $\pm$ 4.50 \\
Random & 81.50 $\pm$ 8.23 & 94.30 $\pm$ 2.80 & 70.75 $\pm$ 4.66 & 22.92 $\pm$ 6.34 & 77.61 $\pm$ 3.46 & 53.73 $\pm$ 4.20 \\
Degree & 81.00 $\pm$ 12.41 & 90.60 $\pm$ 4.88 & 69.65 $\pm$ 5.38 & 25.42 $\pm$ 6.32 & 79.71 $\pm$ 3.40 & 56.34 $\pm$ 4.77 \\
Not using $\ccalT_{\mathrm{miss}}$ & \textbf{84.75 $\pm$ 8.73} & 95.62 $\pm$ 1.56 & 71.28 $\pm$ 4.87 & 21.83 $\pm$ 6.75 & 83.27 $\pm$ 2.13 & 67.23 $\pm$ 5.27 \\ \hline
LSE-NG ($\bar{Q} = 1$) & 83.00 $\pm$ 7.48 & 95.75 $\pm$ 1.65 & 71.68 $\pm$ 3.76 & 24.08 $\pm$ 5.01 & 82.94 $\pm$ 1.90 & 69.02 $\pm$ 4.10 \\
LSE-NG ($\bar{Q} = 3$) & 81.50 $\pm$ 9.89 & 95.17 $\pm$ 1.70 & \textbf{72.79 $\pm$ 4.89} & 24.42 $\pm$ 5.30 & 83.72 $\pm$ 1.37 & 70.28 $\pm$ 2.94 \\ \hline
LSE-NN ($\bar{Q} = 1$) & 81.50 $\pm$ 9.63 & \textbf{96.10 $\pm$ 1.44} & 71.15 $\pm$ 4.57 & 23.83 $\pm$ 6.10 & 83.01 $\pm$ 1.54 & \textbf{71.63 $\pm$ 4.12} \\
LSE-NN ($\bar{Q} = 3$) & 78.25 $\pm$ 8.26 & 95.97 $\pm$ 1.63 & 70.53 $\pm$ 4.07 & \textbf{26.00 $\pm$ 6.13} & \textbf{84.08 $\pm$ 2.10} & 70.33 $\pm$ 3.57 
\vspace{-0.2cm}
    \end{tabular}
    \caption{\small{Accuracy in the test dataset obtained by the baselines and by the approach presented in this work in the downstream task (graph classification). The best performances (excluding those obtained using the true features) are \textbf{bolded}.}}
    \label{tab:results_gclas}
    \vspace{-0.5cm}
\end{table*}

\section{Methodology} \label{sec:methodology}

We introduce our proposed approach to estimate missing node features using structural information and known node features, which is visualized in Fig.~\ref{fig:methodology}.
While the schematic in Fig.~\ref{fig:methodology} illustrates the use of local structure for sharing nodal values, other assumptions can easily be made to associate nodes for feature learning. 

Consider a graph dataset $\ccalT = \{(\ccalG^{(i)},\bbX^{(i)},y^{(i)})\}_{i=1}^T$, 
where for the $i$-th sample we have the graph $\ccalG^{(i)}$ with $N_i$ nodes, $\bbX^{(i)}\in\mathbb{R}^{N_i\times F}$ is a matrix of node features of length $F$, and $y^{(i)}$ is the associated label.
Let $\ccalT_{\mathrm{miss}}\subset \ccalT$ be a subset of $\ccalT$ with missing features, where for every $(\ccalG,\bbX,y)\in\ccalT_{\mathrm{miss}}$, we only know the duplex $(\ccalG,y)$, and define $\ccalT_{\mathrm{full}}=\ccalT\backslash\ccalT_{\mathrm{miss}}$.
Our focus in this work is to recover the missing features $\bbX$ for every $(\ccalG,\bbX,y)\in\ccalT_{\mathrm{miss}}$, resulting in a set $\hat{\ccalT}_{\mathrm{miss}}$ consisting of triplets $(\ccalG,\hat{\bbX},y)$ with approximated features $\hat{\bbX}$.
Subsequently, we may use $\hat{\ccalT}=\ccalT_{\mathrm{full}}\cup\hat{\ccalT}_{\mathrm{miss}}$ for downstream tasks such as graph classification.

Our approach consists of two steps: we first learn a node embedding space preserving graph structural information through which we compute node similarity, and then we predict the values of missing node features using nearby node embeddings.
While we select local structural characteristics as the topological features of interest, note that our framework is amenable to any choice of embedding space that allows us to relate nodes based on similarity.


\medskip

\noindent\textbf{Node embedding space.}
We first obtain a latent space with which we compute node similarity.
We train a GAE to generate node embeddings characterized by node roles, that is, their local structure~\cite{guo20role}.
More precisely, the GAE consists of a graph convolutional network (GCN) as the encoder to learn from structural characteristics, both global and local, and a multilayer perceptron (MLP) as the decoder to invert the embedding process.
For a graph $\ccalG^{(i)}$, the GCN encoder takes as input $\ccalG^{(i)}$ and a corresponding feature matrix $\bbF^{(i)}\in\mathbb{R}^{N_i\times F}$ containing structural information from $F$ features.
We apply the features in~\cite{guo20role}, including local characteristics such as node degree and clustering coefficient, although any features may be used to emphasize different structural behavior.
The parameters $\bbTheta$ of the GAE $f_{\bbTheta}$ are trained to minimize the loss between the output of the GAE and the input feature matrix
\begin{equation*}
    \min_{\bbTheta} \| \bbF^{(i)} - f_{\bbTheta} (\bbF^{(i)}, \ccalG^{(i)})\|_F^2,
\end{equation*}
where $f_{\bbTheta}$ represents the GAE, whose 
output is computed as $f_{\bbTheta} (\bbF, \ccalG^{(i)}) = \mathrm{MLP}_{\bbTheta_2}(\mathrm{GNN}_{\bbTheta_1}(\bbF, \ccalG^{(i)}))$, where $\bbTheta = \{\bbTheta_1, \bbTheta_2 \}$ and the GNN is defined via the following recursion~\cite{kipf17gnns}
\begin{equation}\label{eq:kipf_single_layer}
    \bbH^{(\ell)} = \sigma \left( \tbA \bbH^{(\ell-1)} \bbTheta^{(\ell)} \right),
\end{equation}
where $\bbH^{(\ell)}$ are the hidden features at layer $\ell$; $\bbTheta_1=\{\bbTheta^{(\ell)}\}_{\ell=1}^L$ are the learnable parameters of the $L$ layers; and $\sigma$ is a pointwise non-linearity.
We let $\bbA$ denote the adjacency matrix of the input graph $\ccalG^{(i)}$, and we define $\hat{\bbA}=\bbA+\bbI$, $\hat{\bbD}=\mathrm{diag}(\hat{\bbA}{\bf 1})$, and $\tbA = \hbD^{-1/2} \hbA \hbD^{-1/2}$. Note that the matrix multiplication in \eqref{eq:kipf_single_layer} can be understood as a low-pass graph filtering~\cite{isufi2022graph}, where the nodes average their own value with the values of their neighbors. As a result, the embeddings based on \eqref{eq:kipf_single_layer} will promote similar representations for nodes whose local structural features are similar.
A visualization of the resultant node embeddings from graphs in the molecular classification dataset AIDS~\cite{morris20tudataset} is shown in Fig.~\ref{fig:node_embeddings}.

Given the GAE, we obtain node embeddings for graph $\ccalG^{(i)}$ as $\bbZ^{(i)} = \mathrm{GNN}_{\bbTheta_1} (\bbF^{(i)}, \ccalG^{(i)}) \in \reals^{N_i \times P}$, where $P$ is the dimension of the embedding space.
We further define graph embeddings $\bbz^{(i)}\in\reals^P$ by computing the average across the node dimension, that is, $\bbz^{(i)} = \frac{1}{N_i} \sum_{k=1}^{N_i} \bbz^{(i)}_k \in \reals^{P}$ where $\bbz^{(i)}_k \in \reals^{P}$ is $k$th row of $\bbZ^{(i)}$.

With the node and graph embeddings for every graph in $\ccalT$, we associate $\ccalG^{(i)}\in\ccalT_{\mathrm{miss}}$ with nearby graphs and nodes with respect to the embedding space.
Let $\ccalQ^{(i)}\subset\ccalT_{\mathrm{full}}$ be the set of the $\bar{Q}$ nearest graphs to $\ccalG^{(i)}$ with the same label, that is, for every $\ccalG^{(j)}\in\ccalQ^{(i)}$, we have that $y^{(i)}=y^{(j)}$.
More specifically, all the graphs in $\ccalQ^{(i)}$ are closer to $\ccalG^{(i)}$ than those with the same label but not in $\ccalQ^{(i)}$ using as distance the error between their embeddings $\|\bbz^{(i)}-\bbz^{(j)}\|_2$.
For nearby nodes, we similarly let $\ccalN^{(i,j)}_n$ be the set containing the $\bar{N}$ nodes of graph $\ccalG^{(j)}\in\ccalQ^{(i)}$ closest to node $n$ of graph $\ccalG^{(i)}\in\ccalT_{\mathrm{miss}}$.
That is, all of the $\bar{N}$ nodes from graph $\ccalG^{(j)}$ in $\ccalN^{(i,j)}_n$ are closer (in terms of the same distance previously defined) to node $n$ in graph $\ccalG^{(i)}$ than those not in $\ccalN^{(i,j)}_n$.

\medskip

\noindent\textbf{Predicting missing graph signals.}
Once we are able to compute node similarity, we predict the missing node features using the nearest graphs and nodes.
We propose to generate the features $\hbX^{(i)}$ of $\ccalG^{(i)}\in\ccalT_{\mathrm{miss}}$ as the average of the features in the closest nodes and graphs to $\ccalG^{(i)}$ in $\ccalT_{\mathrm{full}}$ with respect to their embeddings.
Let $\bbC^{(i,j)} \in \reals^{N_i \times N_j}$ be the transformation matrix mapping the features from $\ccalG^{(j)} \in \ccalQ^{(i)}$ to $\ccalG^{(i)}$, where the entry at the $n$th row and $\ell$th column is
\begin{equation}
    C^{(i,j)}_{n,\ell} = \left\{
    \begin{array}{lc}
        \frac{1}{|\ccalN_n^{(i,j)}|} & n \in \{1,...,N_i\}, \ell \in \ccalN_n^{(i,j)} \\ 
        0 & \mathrm{otherwise }
    \end{array} \right.
\end{equation}
For each node $n$ in $\ccalG^{(i)}$, the product $\bbC^{(i,j)}\bbX^{(j)}$ computes the average of the features from the nodes in $\ccalN_n^{(i,j)}$, that is, the closest nodes to $n$ from $\ccalG^{(j)}$. 
Given the set of closest graphs $\ccalQ^{(i)}$ to $\ccalG^{(i)}$, we compute our node feature estimates for $\ccalG^{(i)}$ as 
\begin{equation}\label{eq:feat_est}
    \hbX^{(i)} = \frac{1}{\bar{Q}} \sum_{j \in \ccalQ^{(i)}} \bbC^{(i,j)} \bbX^{(j)}.
\end{equation}
These features complete the triplet $(\ccalG^{(i)}, \hat{\bbX}^{(i)}, y^{(i)})$ for every estimate in $\hat{\ccalT}_{\mathrm{miss}}$, allowing us to use these graphs for downstream tasks such as graph classification. 

\vspace{-0.2cm}

\section{Results}

We showcase the capabilities of our proposed approach for missing feature generation. 
We demonstrate our method in comparison with several baselines in numerical experiments, both for node feature learning and downstream graph classification.

\vspace{0.1cm}

\noindent \textbf{Datasets.} 
We use six real-world data benchmarks: MUTAG, AIDS, PROTEINS and ENZYMES from the TUDataset collection~\cite{morris20tudataset}, and ogbg-molbace and ogbg-molbppp from the OGBG collection~\cite{Hu2020OpenGB}. 
Not only are these standard benchmark datasets for classification tasks, they also provide node features, which allows us to test the hypothesis that node features are relevant for the classification task as well as evaluate our proposed approach. 
The datasets contain either graphs representing molecules (MUTAG, AIDS, ogbg-molbace and ogbg-molbppp), where nodes represent atoms and edges represent chemical bonds, or proteins (PROTEINS) and enzymes (ENZYMES), where nodes represent structural elements and edges encode node proximity.

\vspace{0.1cm}

\noindent \textbf{Experimental setup.} 
We split the dataset with 10\% of the data for validation; 10\% for testing; 30\% for training, or $\ccalT_{\mathrm{full}}$; and 50\% as missing, that is, the set of graphs with missing node features $\ccalT_{\mathrm{miss}}$.
We conduct 15 random realizations of these splits, and the results presented in Tables~\ref{tab:results_npred} and~\ref{tab:results_gclas} list the mean and standard deviation of the metric of interest (to be defined next) across every realization. 

We demonstrate the efficacy of our method on both feature generation and downstream graph classification. 
Our approach that uses the local structure-based node embeddings of nearest neighbors, denoted ``LSE-NN'', is compared to several baselines.
Alternatives to ``LSE-NN'' include classical approaches: ``Degree'' denoting node degree, ``Ones'' for features of all ones, ``Zeros'' for features of all zeros, ``Random'' with features sampled uniformly at random on $[0,1]$.

Our method ``LSE-NN'' exploits not only similar graphs for estimating missing node features but also nodes with similar local structures.
We highlight the benefits of such an approach by also comparing it to a modification denoted ``LSE-NG''.
For this variant, we obtain the nearest graphs $\ccalQ^{(i)}$ for $\ccalG^{(i)}$ as for ``LSE-NN'', but we then assign feature values to each node $k$ in $\ccalG^{(i)}$ by taking node features uniformly at random from the nodes of graphs $\ccalG^{(j)}\in\ccalQ^{(i)}$.
This is equivalent to replacing $\bbC^{(i,j)}$ in \eqref{eq:feat_est} with a random permutation matrix $\bbP^{(i,j)}\in\{0,1\}^{N_i\times N_j}$.
Thus, ``LSE-NG'' predicts node features using similar graphs but does not align nodes by local structure.

%

\vspace{0.1cm}

\noindent \textbf{Feature generation performance.}
We first compare the ability of each method to recover the original node features.
The results presented in Table~\ref{tab:results_npred} show the node feature estimation error as $\|\hat{\bbX}-\bbX\|_F^2/\|\bbX\|_F^2$, where $\hat{\bbX}$ denotes the estimated features and $\bbX$ the true ones.
We see that our architecture consistently beats the alternatives, achieving a lower error in every dataset considered, in some cases by a large margin. 
This shows that the true node features, which are assumed to be the optimal features for downstream tasks, are best recovered by our architecture, without the need for partial observations on the graphs with missing features.

\vspace{0.1cm}

\noindent \textbf{Graph classification performance.}
We also assess the utility of the estimated features for graph classification. 
The results are shown in Table~\ref{tab:results_gclas}, where we present label prediction accuracy using a GNN model trained with the estimated features.
We choose the Graph Isomorphism Network (GIN)~\cite{xu2018gnns} as our GNN architecture, a standard model for graph classification.
For this task, we also add an additional baseline ``Not using $\ccalT_{\mathrm{miss}}$'', where we train the GIN only on the subset of graphs $\ccalT_{\mathrm{full}}$ with known node features, ignoring the graphs in $\ccalT_{\mathrm{miss}}$ with missing features.
In all cases but the MUTAG dataset, the best performance is achieved using the learned GAE embeddings, either from randomly copying nodes from the nearest graphs ``LSE-NG'', which enjoys the best performance on PROTEINS dataset, or by using the nearest nodes ``LSE-NN'', which obtains superior performance for all other datasets.
Thus, not only do we infer missing node features accurately, but the estimates are sufficiently realistic to bolster classification performance when we do not observe the node features of many graphs.


\section{Conclusion}

In this work, we proposed a framework to recover completely missing node features for a set of graphs.
We implemented this framework for estimating features that are characterized primarily by local graph structure.
To this end, we presented a node embedding space using only local topological features.
The embedding space provided a node similarity metric with which we estimated missing node features using similar nodes from nearby graphs.
Our estimates aid graph classification when features are missing, emphasizing the need for accurate nodal characteristics.
In the future, we will generalize to applications such as graph data augmentation, where we can generate synthetic graphs with realistic node features.
Our work connecting node features and graph structure can bolster the success of graph-based learning by exploiting not only structural information but also values explicitly embedded therein.

\bibliographystyle{ieeetr}
\bibliography{citations}

\end{document}